# Bayesian network learning with cutting planes

James Cussens
Dept of Computer Science & York Centre for Complex Systems Analysis
University of York, Deramore Lane, York, YO10 5GH, UK
jc@cs.york.ac.uk

## Abstract

The problem of learning the structure of Bayesian networks from complete discrete data with a limit on parent set size is considered. Learning is cast explicitly as an optimisation problem where the goal is to find a BN structure which maximises log marginal likelihood (BDe score). Integer programming, specifically the SCIP framework, is used to solve this optimisation problem. Acyclicity constraints are added to the integer program (IP) during solving in the form of *cutting planes*. Finding good cutting planes is the key to the success of the approach—the search for such cutting planes is effected using a sub-IP. Results show that this is a particularly fast method for exact BN learning.

## 1 Introduction

A Bayesian network (BN) encodes conditional independence relations between random variables using an acyclic directed graph (DAG) whose vertices are the random variables. The DAG can be used to 'read off' conditional independence relations thus providing insight into the structure of the joint probability distribution represented by the BN. As a result BNs are a very popular probabilistic model and there is great interest in 'learning' BNs from data (i.e. doing statistical model selection where BNs are the model class).

One standard approach to BN learning is 'search and score'. A score is chosen to represent how well any candidate BN is supported by the observed data (and any prior knowledge) and then a search is conducted with the goal of finding a BN with maximal score. BN learning is therefore an optimisation problem. Unfortunately, as is well-known [3], this optimisation problem is NP-hard for any reasonable score, even if the number of parents for any vertex in the DAG is limited to two.

Given this hardness result most BN learning work has concentrated on heuristic search where there is no guarantee that an optimal BN has been found. However, there is a growing body of work on *exact* BN structure learning. One approach is to use dynamic programming [10, 12, 5] which has successfully been used for exact learning up to around 30 vertices.

In this paper *integer programming* (IP) is used for exact BN learning. IP has already been used by Cussens [7] for the special case of BN *pedigree reconstruction* (see Section 8) and by Jaakkola *et al* [9] for general BN learning with a limit on parent set size. As explained in Section 3 the work presented here is most closely related to that of Jaakkola *et al*. However here IP is effected in a more conventional manner than Jaakkola *et al*, using the SCIP (Solving Constraint Integer Programs) framework [2]. In addition a different approach is taken to searching for good cutting planes.

This paper assumes a basic knowledge of what Bayesian networks are, but is intended to be comprehensible by a reader without much knowledge of integer programming. The paper is organised as follows. After explaining the important characteristics of the BDe score in Section 2, a method for encoding the BN structure learning problem as an IP is given in Section 3. The central contribution of the paper is in Section 4 where the cutting plane method is given. After presenting the results of this method in Section 5, ongoing work by the author and related work by others is discussed in Sections 6 and 7. The paper ends with conclusions and pointers to future work (Section 8).

## 2 Scoring Bayesian networks

As is commonly done, candidate BNs are scored using log marginal likelihood with Dirichlet priors over the BN parameters (BDe score). Let $V$ be the set of vertices in a BN, each corresponding to a random variable and let $D$ be the observed data. Throughout this paper $|V|$ will be abbreviated to $n$. The BDe score for DAG $G$ has the following form:

$$\log P(G|D) = \sum_{u \in V} \text{Score}_u(G) \qquad (1)$$

So the total score is a sum of *local scores* $\text{Score}_u(G)$. Moreover (given fixed data $D$) $\text{Score}_u(G)$ is entirely determined by the parents that $u$ has in $G$.

This decomposition motivates the following preprocessing step: for each variable $u$ and each candidate parent set $W$ for $u$, compute and store the corresponding local score. Denote such a local score by $c(u, W)$. Evidently, such an approach is only feasible if there are not too many candidate parent sets. Here this is achieved in the standard way: by limiting the number of parents a vertex can have.

If $c(u, W) < c(u, W')$ for some vertex $u$, and candidate parent sets $W$ and $W'$ where $W' \subset W$, it follows that $W$ cannot be a parent set for $u$ in an optimal BN— replacing $W$ by $W'$ as parents for $u$ in any BN will increase the score and cannot introduce a cycle. As in other work ([6, 8, 7, 9, 5]) this simple observation is used to prune the set of local scores. (Cowell [5] uses this pruning but does not mention it.)

## 3 Encoding the BN learning problem as an integer program

The BN learning problem is encoded as an integer program using the same variables as in some previous work [6, 7, 9]. For each variable $u$ and candidate parent set $W$ corresponding to a stored local score, a binary variable $I(W \rightarrow u)$ is created. $I(W \rightarrow u) = 1$ iff $W$ are the parents of $u$ in an optimal BN. These variables will be referred to as *family variables*. BN learning can then be cast as the following constrained optimisation problem:

Instantiate the $I(W \rightarrow u)$ to maximise:
$\sum_{u,W} c(u, W) I(W \rightarrow u)$
subject to the $I(W \rightarrow u)$ representing a DAG.

Integer programming is linear programming with integrality constraints on the variables. To use an IP approach it is thus necessary to use linear constraints to ensure that only valid DAGs are feasible. Convexity constraints, stating that each variable has exactly one (perhaps empty) parent set, are easy to express as linear constraints:

$$\forall u : \sum_{W} I(W \rightarrow u) = 1 \qquad (2)$$

With (2) any feasible integer solution represents a digraph, but this digraph may have cycles. The key question is how to effectively rule out cycles. A number of approaches are possible. In [7], Cussens considers two approaches. In the first, auxiliary binary variables $I(u < v)$ are created for each distinct pair of variables. $I(u < v) = 1$ indicates that $u$ comes before $v$ in a topological ordering associated with the optimal BN. In the second approach auxiliary integer-valued variables $\text{gen}(v)$ are created where $\text{gen}(v)$ is the index of $v$ in the topological ordering.

Together with appropriate (linear) constraints linking auxiliary variables to $I(W \rightarrow u)$ family variables, these approaches can rule out cycles and produced good results on the relatively easy problem of pedigree learning (see Section 8). However, Jaakkola et al [9] provide a much tighter class of constraints without the need for auxiliary variables. This class of constraints rests upon the observation that any subset $C$ of vertices in a DAG must contain at least one vertex who has no parent in that subset. These are called *cluster-based constraints* and can be expressed as linear constraints:

$$\forall C \subseteq V : \sum_{u \in C} \sum_{W : W \cap C = \emptyset} I(W \rightarrow u) \geq 1 \qquad (3)$$

In this paper, this constraint is generalised. Any DAG has a topological ordering of the variables such that a vertex's parents must appear earlier in the ordering. Given any subset $C$ of vertices in the DAG, the earliest vertex must have no parent in $C$, the next-earliest can have at most one, the next at most two and so on. It follows that the earliest $k$ vertices all have fewer than $k$ parents in the subset. This leads to a linear constraint for each subset $C$ and each $k = 1, \ldots, |C|$ as shown in (4):

$$\forall C \subseteq V, \forall k, 1 \leq k \leq |C| : \qquad (4)$$
$$\sum_{u \in C} \sum_{W : |W \cap C| < k} I(W \rightarrow u) \geq k$$

Call these *k-cluster-based constraints*. Constraint class (3) is a special case of k-cluster-based constraints when $k = 1$; these will be referred to as *1-cluster-based constraints* henceforth.

A k-cluster-based constraint for cluster $C$ has an associated non-negative *surplus variable* $s_C^k$ which measures how far the LHS is above its lower bound. It is interesting to consider the values of surplus variables for DAGs, i.e. integer solutions. $s_{\{u,v\}}^1 = 0$ if and only if there is a line between $u$ and $v$ in the undirected skeleton. For any three vertices $u$, $v$ and $w$, there is an immorality $u \rightarrow w \leftarrow v$ if and only if $s_{\{u,w\}}^1 = 0$, $s_{\{v,w\}}^1 = 0$, $s_{\{u,v\}}^1 = 1$ and $s_{\{u,v,w\}}^1 = 1$. It follows that the Markov equivalence class for a DAG is determined by the values of its surplus variables for clusters of sizes 2 and 3.

If a cluster constraint is tight for a DAG then it lies on the relevant hyperplane. It is interesting to consider when $k$-cluster-based constraints are tight for different values of $k$ since this gives insight into the geometry of the linear polytope. For example, consider three vertices $u$, $v$ and $w$. In a DAG if the subgraph for $\{u, v, w\}$ has three arrows (the maximum) then both 1- and 2-cluster constraints are tight: $s^1_{\{u,v,w\}} = s^2_{\{u,v,w\}} = 0$. For subgraphs of the form $u \to v \to w$ or $u \leftarrow v \to w$ the 1-cluster is tight but not the 2-cluster. For subgraphs of the form $u \to v \leftarrow w$ (immoralities) the 2-cluster is tight but not the 1-cluster.

## 4 Cutting planes

To rule out cycles it is enough to add all 1-cluster-based constraints, but there are far too many of these to include in an integer program. Instead such constraints will be added as *cutting planes* during the solving process.

Initially an integer program with just the convexity constraints (2) is constructed. Following the standard approach to integer programming, the linear relaxation of this IP is solved. In the linear relaxation the integrality constraint on family variables is removed and they can take any value in $[0, 1]$. The linear relaxation (which is a linear program (LP)) is quickly solved using (a version of) the simplex algorithm, resulting in an *LP solution* $\check{x}$ which is an instantiation of all family variables.

This first LP solution will correspond to choosing the best scoring parents for each vertex and (barring exceptional cases) will contain cycles. It is thus necessary to add a cluster-based constraint to remove the LP solution from the feasible region. This is known as *separation*. This leads to a new IP whose linear relaxation can then also be solved.

A linear constraint added to an IP to remove an LP solution is known as a *cutting plane* since the LP solution is cut away from the feasible region. Note that the feasible region for the linear relaxation will be a convex polytope and LP solutions will be vertices of this polytope. Adding a cutting plane 'snips' off this vertex.

The approach taken here is to repeatedly (1) solve the current linear relaxation and (2) add cutting planes until either no further cutting planes can be found or an LP solution corresponding to a DAG is returned. Since the score of an LP solution is an upper bound on that of the optimal BN any such DAG will be the optimal BN.

### 4.1 Searching for good cutting planes

The key issue is how to quickly find good cutting planes. The overall goal is to reduce the upper bound given by LP solutions as much as possible by cutting deep into the LP polytope. It is generally beneficial to add several cutting planes removing a given LP solution. There are three important quality criteria for a collection of cutting planes "the efficacy of the cuts, i.e., the distance of their corresponding hyperplanes to the current LP solution, the orthogonality of the cuts with respect to each other, and the parallelism of the cuts with respect to the objective function." [1].

In this work the focus is on finding cutting planes with high efficacy (and hoping for the best as regards orthogonality and objective function parallelism). Let $\check{x}_{I(W \to u)}$ be the value of $I(W \to u)$ in the current LP solution $\check{x}$, then if the 1-cluster constraint defined using cluster $C$ is a cutting plane removing $\check{x}$ then its efficacy is:

$$\frac{1 - \sum_{u \in C} \sum_{W: W \cap C = \emptyset} \check{x}_{I(W \to u)}}{\sqrt{\sum_{u \in C} \sum_{W: W \cap C = \emptyset} 1}} \quad (5)$$

Instead of attempting to maximise (5) directly its numerator is maximised by searching for a cluster $C$ minimising

$$\sum_{u \in C} \sum_{W: W \cap C = \emptyset} \check{x}_{I(W \to u)} \quad (6)$$

This minimisation is effected by a sub-IP. In this sub-IP a binary variable $J(W \to u)$ is created for each family variable $I(W \to u)$ in the main IP. The objective function for the sub-IP is:

$$\sum_{u,W} \check{x}_{I(W \to u)} J(W \to u) \quad (7)$$

A constraint is placed stating that only solutions where the objective value is strictly less than 1 are permitted. This ensures that all solutions correspond to a cutting plane. Constraints are then placed upon the $J(W \to u)$ such that any feasible joint instantiation corresponds to a cluster $C$ and where (7) is equivalent to (6) for this cluster. In other words, $J(W \to u) = 1$ iff $u \in C$ and $W \cap C = \emptyset$.

A binary variable $I(u)$ is created for each $u \in V$. $I(u) = 1$ iff $u$ is in the cluster $C$. For any $J(W \to u)$ we then have the following constraint:

$$J(W \to u) = 1 \Leftrightarrow I(u) = 1 \wedge \bigwedge_{u' \in W} I(u') = 0 \quad (8)$$

Note that from (8) it follows that $J(\emptyset \to u) = 1 \Leftrightarrow I(u) = 1$ and so in the actual sub-IP $I(u)$ variables are replaced by $J(\emptyset \rightarrow u)$ variables. The final constraint in the sub-IP is that at least 2 of the $J(\emptyset \rightarrow u)$ are set to 1.

An empirical comparison between a pure constraint processing approach to solving this sub-IP and the standard approach using linear relaxations shows that the former is more efficient. The constraints (8) are effected using SCIP's built-in 'and' constraint. Depth-first search is then used to search for a minimising feasible solution. Only $J(\emptyset \rightarrow u)$ variables are made available for branching. To choose between these, SCIP's default branching score function is replaced by one which attempts to balance the search tree as much as possible. Once a $J(\emptyset \rightarrow u)$ variable has been branched on the $J(\emptyset \rightarrow u) = 1$ subtree is always explored first.

In most cases, the sub-IP was run until an optimal cluster was found, the rationale being that this effort is worth expending to find good cutting planes. SCIP does not just return the optimal solution but all sub-optimal solutions found during the search. Collecting all solutions is important since each corresponds to a cutting plane and the joint benefit of a collection of cutting planes can be considerably greater than one (apparently) best one.

To reduce the number of variables involved in a cutting plane, the sum $\sum_{W:W \cap C = \emptyset} I(W \rightarrow u)$ in (3) is replaced with $1 - \sum_{W:W \cap C \neq \emptyset} I(W \rightarrow u)$ whenever the latter has fewer summands.

When the LP solution happens to be entirely integral and corresponds to a cyclic digraph the sub-IP is solved very quickly and only one cutting plane is returned. This is because the sub-IP is just finding a cycle in a manner similar to that of standard depth-first search [4]. When the LP solution contains fractional values, solving is slower but multiple cutting planes are found, sometimes more than 100.

Only $k$-cluster-based constraints for $k = 1$ are searched for by this sub-IP. It is of course possible to set up further sub-IPs to search for $k$-cluster-based constraints for $k > 1$. This was done for $k = 2$ but the time taken to solve this sub-IP was not compensated by the effectiveness of the cutting planes found. Instead a simple approach is taken: for each 1-cluster-based constraint found the corresponding 2-cluster-based constraint is also added.

### 4.2 Gomory cuts

Using the DAG-specific cutting planes described above is the key to the success of this IP approach to BN learning. However, these are not the only useful cutting planes. It can happen that there are no 1-cluster-based constraints separating the current LP solution. If that is the case then SCIP is asked to look for *Gomory cuts*. Gomory cuts are general-purpose cutting planes which can be quickly computed from the simplex tableau. For further information on Gomory cuts see, for example, Wolsey [13]. Relying on Gomory cuts alone would be inefficient since they are generally not very deep. However, they turn out to be crucial (see Section 5). Even a weak Gomory cut at least separates the current LP solution thus leading to a new linear relaxation—one with a different LP solution, for which the cutting plane algorithm described in Section 4.1 may well find good cutting planes. In short, Gomory cuts can help a pure DAG-specific cutting plane approach from getting stuck. However, it has been found important to only add Gomory cuts when no cluster-based constraints can be found (*delayed Gomory cuts*). Since quite weak cuts are included in the approach presented here, adding all possible Gomory cuts can lead to poor results. For example, as will be shown in Section 5, the Water100 problem can be solved in 9 seconds using 2196 generated cuts with a delayed Gomory cut approach. With Gomory cuts not delayed solving was abandoned after more than 5.5 hours of failing to find an optimal BN! In this case the large number of cuts generated (over 6745) led to very slow LP solving with over 1.3 million LP iterations done by the time of abandonment.

Gomory cuts have another, psychological, benefit. By inspecting Gomory cuts and determining what they mean in graph-theoretic terms it is possible to discover new classes of cutting planes. This is how it came to be realised that the 1-cluster-based constraint of Jaakkola *et al* could be generalised to $k$-cluster-based constraints.

## 5 Results

The approach described in the preceding two sections was implemented in C using the SCIP framework. SCIP is available from `http://scip.zib.de/`. SCIP has a plug-in architecture and the approach presented here was effected by writing a *constraint handler*. (See `http://scip.zib.de/doc/html/CONS.html`.) CPLEX 12 was used as the LP solver. All experiments were performed using a 64-bit Linux kernel on a single-thread of a dual 3GHz CPU with 3.8Gb of RAM. SCIP was configured to allow the inclusion of even very weak cuts. For Gomory cuts the relevant C source in SCIP was edited to allow very low quality Gomory cuts to be generated. (See Section 4.1 for an explanation of cut quality.)

Ten different BNs were used. Various datasets were sampled from them and local scores were computed. The details can be found in Table 1. The 'upper' collection of BNs in Table 1 all came from Bayesian Network Repository (http://compbio.cs.huji.ac.il/Repository/). Local scores were computed using an unoptimised Python program which meant that it could take as long as an hour to compute local scores for the 3 slowest cases, with the most slow taking a little over 6 hours. The 'lower' 4 datasets were supplied as local scores and are the same as those used in [9].

| Name | $n$ | $m$ | Families |
|---|---|---|---|
| Mildew100 | 35 | 3 | 3513 |
| Mildew1000 | 35 | 3 | 161 |
| Mildew10000 | 35 | 3 | 463 |
| Water100 | 32 | 3 | 482 |
| Water1000 | 32 | 3 | 573 |
| Water10000 | 32 | 3 | 961 |
| alarm100 | 37 | 3 | 907 |
| alarm1000 | 37 | 3 | 1928 |
| alarm10000 | 37 | 3 | 6473 |
| asia100 | 8 | 3 | 41 |
| asia1000 | 8 | 3 | 107 |
| asia10000 | 8 | 3 | 161 |
| carpo100 | 60 | 3 | 5068 |
| carpo1000 | 60 | 3 | 3827 |
| carpo10000 | 60 | 3 | 16391 |
| hailfinder100 | 56 | 3 | 244 |
| hailfinder1000 | 56 | 3 | 761 |
| hailfinder10000 | 56 | 3 | 3768 |
| insurance100 | 27 | 3 | 279 |
| insurance1000 | 27 | 3 | 774 |
| insurance10000 | 27 | 3 | 3652 |
| alarm | 37 | 4 | 5630 |
| mirna | 22 | 4 | 1279 |
| phen | 25 | 4 | 2296 |
| wdbc | 31 | 4 | 5921 |

Table 1: Datasets, parent set size limits and family variables used in experiments. $n$ is the number of variables in the dataset, $m$ is the limit on parent set size and 'Families' is the number of family variables remaining after pruning. For the upper set of datasets the number in the dataset's name records the number of datapoints sampled from the relevant BN. For alarm, mirna, phen and wdbc the number of datapoints sampled was 1000, 218, 926 and 569 respectively.

The main results of this paper are contained in the 'Time' columns of Tables 2 and 3 which display the time in seconds (rounded to the nearest second, as measured by the UNIX time command, system call time included) to find an optimal BN for all datasets. In Table 2 both 1- and 2-cluster constraints are added. In Table 3 only 1-cluster constraints are used and results for 'easy' problems have been omitted in the interests of space—they are similar to those found in Table 2. Neither option is consistently better, however using 1-cluster constraints only solves the biggest problem (carpo10000) in $\approx$ 7.4 hours whereas with both 1- and 2-cluster constraints it takes almost 12 hours. It seems likely that a more careful combination of 1- and 2- cluster constraint cutting planes will be the best option.

For two datasets, carpo100 and carpo10000, it was necessary to impose a time limit of 100 seconds on the sub-IP used to find cutting planes: any solutions found within this limit are added as cutting planes.

On a number of datasets SCIP reached a point at which it could find no further cutting planes but had not found an optimal BN and thus had to resort to branch-and-bound. The 'Nodes' column in the tables is the number of nodes in the branch-and-bound tree at the time of solution, so that if Nodes=1 the problem was solved without branching. A *branch-and-cut* approach was taken: cluster-based cutting planes and Gomory cutting planes are generated not only in the root node of the branch-and-bound tree but at *every other node in the tree*.

SCIP has a collection of *primal heuristics* which periodically search for good solutions. Even if such a solution does not turn out to be optimal it can be useful since it may allow pruning of the branch-and-bound tree. In this paper only fast primal heuristics were allowed to run. Typically, the optimal solution (and a number of suboptimal solutions) were found during the running of the simplex algorithm computing a LP solution: if an integral solution is visited in one of the simplex iterations and it turns out to be a valid DAG with a better score than the current incumbent solution, it becomes the new incumbent. This is an instance of SCIP's *simple rounding* primal heuristic; in this case there are zero roundings required to produce a solution. In some cases a more sophisticated rounding approach which uses the LP solution, found the optimal BN. Turning off all heuristics modestly speeds up solving for all problems which can be solved in the root, since in that case it is enough to wait for the cutting planes to lead to an LP solution which is entirely integral.

In Tables 2 and 3 'Cuts' is the total number of cuts generated whereas 'Rows' is the total number of linear constraints in effect at the point at which an optimal solution was found; this number is only really informative for problems solved without branching. 'Rows' is generally smaller than 'Cuts' since cuts can become redundant with the addition of further cuts. The number of cuts is rounded to the nearest 1000 where the number is large.

| Name | Time | Nodes | Rows | Cuts |
|---|---|---|---|---|
| Mildew100 | 4 | 1 | 118 | 121 |
| Mildew1000 | 1 | 1 | 97 | 86 |
| Mildew10000 | 3 | 1 | 1372 | 1630 |
| Water100 | 9 | 1 | 422 | 2196 |
| Water1000 | 5 | 1 | 615 | 1855 |
| Water10000 | 20 | 25 | 228 | 2729 |
| alarm100 | 4 | 1 | 260 | 582 |
| alarm1000 | 15 | 19 | 211 | 1324 |
| alarm10000 | 12872 | 23762 | 345 | 84000 |
| asia100 | 0 | 1 | 29 | 13 |
| asia1000 | 0 | 1 | 75 | 77 |
| asia10000 | 7 | 15 | 50 | 5287 |
| carpo100(*) | 15178 | 226 | 433 | 34000 |
| carpo1000 | 593 | 116 | 445 | 10000 |
| carpo10000(*) | 42275 | 4574 | 802 | 50000 |
| hailfinder100 | 1 | 1 | 125 | 99 |
| hailfinder1000 | 5 | 1 | 305 | 955 |
| hailfinder10000 | 169 | 177 | 267 | 3903 |
| insurance100 | 1 | 1 | 150 | 194 |
| insurance1000 | 3 | 1 | 238 | 624 |
| insurance10000 | 54 | 61 | 196 | 2033 |
| alarm | 153 | 62 | 211 | 2084 |
| mirna | 5 | 1 | 673 | 1446 |
| phen | 9 | 7 | 119 | 774 |
| wdbc | 78 | 1 | 1105 | 1995 |

Table 2: Time taken to find optimal BNs using both 1- and 2-cluster constraints. Problems marked (*) had a 100 second time limit on the sub-IP. For details see the main body of the text.

| Name | Time | Nodes | Rows | Cuts |
|---|---|---|---|---|
| Water1000 | 115 | 1 | 604 | 2121 |
| Water10000 | 28 | 1 | 507 | 2202 |
| alarm1000 | 20 | 28 | 208 | 1185 |
| alarm10000 | 7056 | 3109 | 362 | 6651 |
| asia10000 | 11 | 8 | 49 | 5788 |
| carpo100(*) | 27860 | 373 | 644 | 29000 |
| carpo1000 | 434 | 132 | 589 | 5451 |
| carpo10000(*) | 26774 | 5205 | 796 | 18000 |
| hailfinder10000 | 934 | 982 | 348 | 5663 |
| insurance10000 | 61 | 56 | 214 | 1389 |
| alarm | 97 | 7 | 189 | 929 |
| phen | 12 | 13 | 142 | 559 |
| wdbc | 68 | 1 | 651 | 1057 |

Table 3: Time taken to find optimal BNs using only 1-cluster constraints. Problems marked (*) had a 100 second time limit on the sub-IP. For details see the main body of the text.

To test the effectiveness of Gomory cuts, solving was attempted again for each dataset with both 1- and 2-cluster constraints but with Gomory cuts disabled. For all but the largest problems solving was still successful and was generally a little quicker. However for big problems disabling Gomory cuts could be disastrous. alarm10000 took 18462 seconds to solve (as opposed to 12872 with Gomory cuts). hailfinder10000, which took only 169 seconds to solve with Gomory cuts had still failed to find an optimal BN after 85075 seconds without them (and had exhausted RAM by this point). With Gomory cuts, the $358 - 56 = 302$ non-redundant cuts generated for hailfinder10000 had reduced the upper bound on the objective function to $-4.970794 \times 10^5$ before SCIP had to resort to branching. Since the optimal BN has score $-4.976318 \times 10^5$ this upper bound is quite tight allowing for a reasonably fast branch-and-cut search. Without Gomory cuts the $502 - 56 = 446$ non-redundant cuts generated had only reduced this upper bound to $-4.946169 \times 10^5$. Using Gomory cuts had allowed for fewer but better cuts. The very largest problems, carpo100 and carpo10000 were not even attempted without Gomory cuts.

### 5.1 How to reproduce these results

Go to:
http://www.cs.york.ac.uk/~jc/research/uai11

## 6 Ongoing work

The approach described above is the most successful to date but there are a number of elaborations which have been attempted. Although these have yet to render any consistent improvement it is instructive to list them since (1) it may be that some variant will succeed and (2) negative results are of independent interest.

The most obvious idea is to add a reasonably small number of $k$-cluster constraints directly into the IP so that they do not have to be found as cutting planes. There is the possibility that SCIP's pre-processing algorithms might be able to simplify the problem using these constraints. In the event, pre-adding $k$-cluster constraints led to slower solving and increased memory requirements.

Since SCIP has to resort to branching on bigger problems a *primal heuristic* specific to BN learning is motivated. This is a function, with access to any current LP solution, which generates DAGs in the hope that they score better than the current incumbent solution. High scoring incumbent solutions allow pruning of the branch-and-cut tree thus speeding up solving, sometimes considerably. Two primal heuristics have been attempted. The first very simple approach is to gener-ate very many random orders of the BN variables and return the highest-scoring BN consistent with each order, ultimately returning the best BN overall. In the second a variable ordering is generated from the current LP solution with a view to being 'as consistent as possible' with it. The best BN for that ordering is then proposed. Both of these approaches are quick, but the BNs found are less useful than those found by rounding (Section 5).

The presented approach makes no use of *Markov equivalence classes* which partition the BN space into equivalent statistical models. BNs in a Markov equivalent class will have the same score, for any reasonable scoring function. Since two BNs are Markov equivalent if they have the same undirected skeleton and set of 'immoralities' (unmarried parents) $\binom{n}{2}$ auxiliary binary *line variables* were created each indicating the presence/absence of an edge in the undirected skeleton. Since such variables can be defined in terms of family variables there is no real increase in the number of variables in the IP. Branching priorities were set so that branching could only occur on line variables. The idea here is that branching divides Markov equivalence classes rather than just BNs. In most cases this approach slowed down solving but for alarm10000 and alarm solving took only 6526 and 84 seconds respectively, which is better than the results reported in Tables 2 and 3.

## 7 Related work

As previously noted, this work is most closely related to that of Jaakkola *et al* [9], since the key 1-cluster constraints were introduced in that paper. The main difference is that these constraints are applied here in a conventional manner: as cutting planes in an existing integer programming framework (SCIP). Although, as here, Jaakkola *et al* add 1-cluster constraints 'on the fly' they use a different search approach to find good clusters (actually two search methods are used). When branching they always choose the node with the best LP solution (in the hope that the optimal BN is in that subtree). The variable to branch on is determined by a rule informed by the existence of cluster constraints. Here, in contrast, SCIP's default approach to branch-and-cut is used.

An empirical comparison is also possible since the same local scores have been used. Jaakkola *et al* used a 2.4GHz Dual Core Macbook pro Laptop with 4GB of memory and report solving times of 8 seconds, 7 minutes and 21 minutes for datasets phen, wdbc and alarm respectively. Here these were solved in 9 seconds, 1.3 minutes and 2.55 minutes respectively. Note that SCIP does not resort to branching on 2 of the 4 datasets from Jaakkola *et al*.

Other much bigger problems have also been solved, albeit sometimes very slowly. Note that in common with Jaakkola *et al* no consideration is given to how 'close' the found optimal BNs are to the true data-generating BNs, the focus here is exclusively on the 'search' part of a 'search-and-score' learning algorithm. This closeness depends crucially on the score (particularly the choice of Dirichlet priors [11]) as well as, of course, the amount of data.

## 8 Conclusions and future work

The main finding of this paper is that integer programming can be used for exact learning of Bayesian networks with quite a large number of variables, the carpo datasets have 60 BN variables—as long as there is a limit on parent set sizes.

In some cases there is a known limit on parent set sizes. For example in the BN representation of a *pedigree*, variables represent individuals and the arrows in the DAG represent real parenthood relationships. Pedigrees are often called 'family trees' although the DAG need not be a tree if there has been inbreeding. Pedigrees are learnt (or 'reconstructed' as it is known in the statistical genetics literature) from DNA marker data. Cowell [5] has applied a dynamic programming approach to exact pedigree reconstruction whereas Cussens [7] used the IP approach outlined in Section 3

In most other applications the limit on parent set size is artificial. An interesting avenue for future work is to see whether a *column generation* approach using a *variable pricer* can be used to overcome this limitation. A variable pricer allows the dynamic generation of new variables. Here these would be new family variables. SCIP provides a convenient and effective way of creating variable pricers due to its plugin approach.

A more fundamental issue is whether it is worth the effort to do 'exact' BN learning to select a single BN model as our best guess for the true BN. One motivation is that an optimal BN can be very much more probable than any sub-optimal one—the relevant landscape is very 'spiky', particularly for large datasets. (Recall that BDe corresponds to posterior probability with a uniform structure prior.) However, from a Bayesian perspective at least, returning only a single model is dubious since it gives little notion of the degree of model uncertainty inherent in any learning problem. However, the current approach can be used as a subroutine in a search-based model averaging approach (but without pruning of local scores). After the optimal BN has been found its entire Markov equiva-lence class can be ruled out via appropriate constraints and a new search can be started. An advantage of this approach is that there is a guarantee that any yet-to-be-found BNs can have a score no better than those found already. This could be used to provide a bound of the probability mass found so far.

**Acknowledgements**

Many thanks to Tommi Jaakkola for providing datasets and answering various questions. Thanks to Ricardo Silva for pointing me to Jaakkola *et al* [9]. The following SCIP developers helped with various aspects of their framework: Tobias Achterberg, Timo Berthold, Ambros Gleixner, Stefan Heinz, Michael Winkler and Kati Wolter. Particular thanks to Kati Wolter for advice on how to generate extra Gomory cuts. Three anonymous reviewers provided valuable advice on how to improve the original submission.